\pdfoutput=1

\documentclass[11pt]{article}

\usepackage{acl}

\usepackage{graphicx}
\usepackage{booktabs}
\usepackage{amsmath}
\usepackage{amssymb}

\usepackage{times}
\usepackage{latexsym}

\usepackage[T1]{fontenc}

\usepackage[utf8]{inputenc}

\usepackage{microtype}

\renewcommand{\vec}[1]{\mathbf{#1}}

\newcommand*{\lsem}{\ensuremath{\mathcal{L}_\mathsf{SemantiX}}}
\newcommand*{\wce}{Whisper ($\mathcal{L}_\text{CE}$) }
\newcommand*{\wsem}{Whisper ($\mathcal{L}_\mathsf{SemantiX}$) }

\title{Spaiche: Extending State-of-the-Art ASR Models to Swiss German Dialects}

\author{Clément Sicard \\
  ETH Zürich \\
  \texttt{csicard@ethz.ch} \\\And
  Victor Gillioz \\
  ETH Zürich \\
  \texttt{vgillioz@ethz.ch} \\
  \And
  Kajetan Pyszkowski \\
  ETH Zürich \\
  \texttt{kpyszkowski@ethz.ch} \\}

\begin{document}
\maketitle
\begin{abstract}
Recent breakthroughs in Natural Language Processing (NLP) largely increased the presence of ASR systems in our daily lives. However, for many low-resource languages, ASR models still need to be improved due in part to the difficulty of acquiring pertinent data. This project aims to help advance research in ASR models for Swiss German dialects, by providing insights about the performance of state-of-the-art ASR models on recently published Swiss German speech datasets.

We propose a novel loss that takes into account the semantic distance between the predicted and the ground-truth labels. We outperform current state-of-the-art results by fine-tuning OpenAI's Whisper model on Swiss German datasets.
\end{abstract}

\section{Introduction}
Swiss German dialects are spoken by around $5$ million people in Switzerland and are used in day-to-day life as a primary language of communication in the country. However, due to the non-existence of a standardized written form, standard German is used as the primary form of written communication outside of informal text messages. For this reason, we focus on transcription from multi-dialect Swiss German speech to standard German text, for which an extensive offer of language processing tools already exists.

This paper presents the results of evaluations we conducted on the recently available Swiss German datasets SwissDial \cite{swissdial}, SDS-200 \cite{sds-200}, and SPC \cite{spc}. Two models were used to conduct these evaluations: XLS-R \cite{xls-r}, and Whisper \cite{whipser} models. Both were evaluated and fine-tuned on Swiss German audio data with standard German annotations following transfer learning training strategies to respond to the low-resource situation of the Swiss German language.

Based on recent discussions\footnote{\url{https://www.speechmatics.com/company/articles-and-news/the-future-of-word-error-rate}} about the relevance of Word Error Rate (WER) \cite{wer}, we additionally implement a semantic distance metric for evaluation by making use of language model embedding. We also use this metric to define a custom training loss to fine-tune Whisper model.

Our experiments show encouraging results for Whisper model. They suggest a soon-to-come availability of Swiss German ASR services, in a significant part due to the publication of annotated Swiss German speech datasets since 2021. We observe impressive results in Zero-Shot evaluations of Whisper models, and it even outperforms state-of-the-art results after training on a limited number of epochs. To the best of our knowledge, on the \textit{multi-speaker-multi-dialect} SDS-200 dataset, we produce the current best results, with a WER of $20.6$ and a BLEU of $66.6$. We also achieve a better BLEU on the SPC dataset, namely $61.6$. The model also offers high complexity in its outputs by including capital letters, numbers and punctuation with high accuracy.
\section{Models \& Methods}

During our whole experiment, we extensively used HuggingFace to host our datasets, training results and models in different training stages. We also used multiple HuggingFace libraries to share datasets and train on multiple GPUs.

\begin{table*}[ht]
\small
\begin{center}
    \begin{tabular}{l*{4}{|*{3}{c}}}
    \toprule
    \multicolumn{1}{c}{} &
    \multicolumn{3}{c}{\textbf{SwissDial}} &
    \multicolumn{3}{c}{\textbf{SDS-200}} &
    \multicolumn{3}{c}{\textbf{SPC}} &
    \multicolumn{3}{c}{\textbf{Fleurs}} \\
    Models &
    {WER} & {CER} & {BLEU} &
    {WER} & {CER} & {BLEU} &
    {WER} & {CER} & {BLEU} &
    {WER} & {CER} & {BLEU} \\
    \midrule
    XLS-R 1B
    & $69.3$ & $33.8$ & $12.3$
    & $76.6$ & $40.4$ & $8.8$
    & $73.6$ & $39.9$ & $12.7$
    & $10.2$ & $5.6$ & $71.3$ \\
    \midrule
    Whisper Tiny
    & $80.2$ & $42.3$ & $10.3$
    & $92.7$ & $54.9$ & $6.61$
    & $96.3$ & $55.7$ & $9.7$
    & $37.5$ & $12.9$ & $46.0$ \\
    Whisper Base
    & $66.8$ & $33.2$ & $19.0$
    & $78.0$ & $43.4$ & $13.8$
    & $70.4$ & $37.8$ & $20.3$
    & $25.6$ & $8.4$ & $60.2$ \\
    Whisper Small
    & $46.8$ & $22.6$ & $36.4$
    & $51.0$ & $27.3$ & $34.2$
    & $50.9$ & $27.1$ & $37.3$
    & $14.7$ & $4.1$ & $75.0$ \\
    Whisper Medium
    & $33.8$ & $16.9$ & $50.8$
    & $36.8$ & $20.4$ & $49.7$
    & $37.4$ & $20.2$ & $50.8$
    & $10.5$ & $2.9$ & $81.7$ \\
    Whisper Large
    & $\mathbf{29.4}$ & $\mathbf{14.8}$ & $\mathbf{56.2}$
    & $\mathbf{31.7}$ & $\mathbf{18.0}$ & $\mathbf{55.6}$
    & $\mathbf{33.2}$ & $\mathbf{18.2}$ & $\mathbf{55.6}$
    & $\mathbf{8.7}$ & $\mathbf{2.3}$ & $\mathbf{84.7}$ \\
    \bottomrule
    \end{tabular}
    \caption{Baseline performances on a Zero-Shot evaluation \label{zeroshot}}
\end{center}
\end{table*}

\label{datasets}
\subsection{Datasets}

We conducted a series of experiments for this paper with the four following datasets. Due to their respective particularities, they were used to different ends.

\subsubsection{SwissDial}

The SwissDial \cite{swissdial} dataset is an annotated parallel corpus of spoken Swiss German across eight major dialects (AG, BE, BS, GR, LU, SG, VS, ZH) with Swiss German and High German transcripts. 
It includes around 3 hours of high-quality audio per dialect. However, the dataset contains a class imbalance as it has around three times more Grisons Swiss German than any other dialect, which must be considered when training and evaluating models.

We randomly selected $20\%$ of the data to constitute a test set. However, it must be noted that there is a single speaker per dialect, and most of the sentences are spoken across the eight dialects – for this reason, the independence of the train and test set is not perfect, and over-optimistic evaluation results are expected.

\subsubsection{Swiss Parlament Corpus}

The Swiss Parliament Corpus (SPC) \cite{spc} is a dataset of transcriptions of parliamentary speeches and proceedings from the Swiss National Council and Council of States. It consists of 293 hours of data. The corpus contains automatically aligned transcripts of speeches and proceedings from various parliamentary sessions, mainly in the dialect from Bern canton, with transcriptions in standard German. However, because of its size and the fact that there are almost only data points from Bern, in a juridic context, the usage of this dataset may introduce a bias in the training of a multi-dialect Swiss German ASR model. Moreover, the audio samples are often noisy, which brings additional difficulty for transcription – although Whisper is proved to be robust on noisy examples.

\subsubsection{SDS-200}

SDS-200 \cite{sds-200} dataset consists of $200$ hours of speech data in Swiss German and corresponding transcriptions in standard German. The speech data was recorded from approximately $4$k native speakers of Swiss German and covered various topics and Swiss German dialects. The speech data was recorded with a web tool open to the public. It covers a large part of the Swiss German dialect landscape. The dialect distribution roughly follows the speaker distribution in Switzerland.

The samples have been partly validated by the public and is mostly read-aloud data. Moreover, the test set only contains speakers who are not present in the train set, and for which the speech audio has been sufficiently validated and is judged of high quality by the authors. We, therefore, consider it an essential set for model result evaluation.

\subsubsection{Fleurs}
Fleurs \cite{Fleurs} is a speech dataset by Google which contains samples in $102$ different languages, with approximately $12$ hours of speech supervision per language. The dataset was built on top of the machine translation FLoRes-101 benchmark \cite{Flores-101}. We use the standard German part of the dataset for evaluation.

\subsubsection{ArchiMob}

Another commonly referenced Swiss German audio dataset is ArchiMob \cite{samardzic-etal-2016-archimob}. It consists of $43$ interview recordings in $14$ different Swiss German dialects for around $70$ hours of data. However, we decided against using this dataset in this project as it does not offer standard German transcription.

\begin{table*}[ht]
\small
\begin{center}
    \begin{tabular}{l*{4}{|*{3}{c}}}
    \toprule
    \multicolumn{1}{c}{} &
    \multicolumn{3}{c}{\textbf{SwissDial}} &
    \multicolumn{3}{c}{\textbf{SDS-200}} &
    \multicolumn{3}{c}{\textbf{SPC}} &
    \multicolumn{3}{c}{\textbf{Fleurs}} \\
    Models &
    {WER} & {CER} & {BLEU} &
    {WER} & {CER} & {BLEU} &
    {WER} & {CER} & {BLEU} &
    {WER} & {CER} & {BLEU} \\
    \midrule
    \citealp{sds-200} (norm.)
    & $-$ & $-$ & $-$
    & $21.6$ & $-$ & $64.0$
    & $-$ & $-$ & $-$
    & $-$ & $-$ & $-$ \\
    \citealp{most-recent-sg-asr} (norm.)
    & $-$ & $-$ & $-$
    & $-$ & $-$ & $-$
    & $\mathbf{23.7}$ & $-$ & $60.7$
    & $-$ & $-$ & $-$ \\
    \midrule
    \midrule
    XLS-R 1B
    & $17.7$ & $10.3$ & $62.9$
    & $25.2$ & $16.4$ & $53.6$
    & $40.2$ & $24.0$ & $37.9$
    & $47.9$ & $25.8$ & $25.0$ \\
    \midrule
    \wce
    & $\mathbf{14.3}$ & $\mathbf{7.5}$ & $\mathbf{77.7}$
    & $21.2$ & $\mathbf{12.9}$ & $65.4$
    & $28.9$ & $\mathbf{16.0}$ & $\mathbf{61.6}$
    & $\mathbf{16.6}$ & $\mathbf{5.4}$ & $72.9$ \\
    \wsem
    & $21.2$ & $9.8$ & $65.3$
    & $\mathbf{20.6}$ & $13.0$ & $\mathbf{66.6}$
    & $30.4$ & $18.6$ & $56.7$
    & $\mathbf{16.6}$ & $5.5$ & $\mathbf{73.0}$ \\
    \bottomrule
    \end{tabular}
    \caption{Performance of our models on a post-training evaluation \label{post-training}}
\end{center}
\end{table*}

\subsection{Models}

In this project, we establish a baseline using Facebook's XLS-R 1B \cite{xls-r} and further train OpenAI's Whisper \cite{whipser} as our primary focus. As Whisper's medium version has $769$M parameters and its large version has $1.5$B, we believe that the comparison between the two architectures is reasonable. Furthermore, to compute our custom loss \lsem, which requires the use of a LLM, we use a pre-trained XLM-RoBERTa \cite{roberta} for multilingual sentence embeddings \cite{LLM-loss}. However, as this model is not the main focus of this research, we do not present it in detail here.

\subsubsection{Facebook's XLS-R 1B}

XLS-R \cite{xls-r} is a multilingual speech representation learning model. It is based on Wav2Vec2 \cite{wav2vec2} self-supervised learning framework and was pre-trained on $436$k of unlabelled speech from 128 languages. No Swiss German data was used for training. The most recent publications on Swiss German ASR \cite{sds-200} \cite{most-recent-sg-asr} showed XLS-R capacity for Swiss German speech to standard German transcription. Following these publications, we similarly focused on the 1B parameters model (XLS-R 1B).

We implemented this model to offer a basis for comparison to Whisper results when trained in a similar setting. Moreover, results on the SwissDial dataset have yet to be published. For this reason, the training and evaluation of the XLS-R 1B model allowed us to set a first benchmark and provide a better perspective on our results.

To respond to the low-resource setting of Swiss German and the limited scope of this research, we use a model that has already been fine-tuned on multiple German speech datasets \cite{grosman2021xlsr-1b-german}. It also offers a CTC \cite{ctc} beam search decoder language model, which improved our overall results. The output vocabulary consists of lowercase letters and a limited number of special characters. This model reached a WER \cite{wer} of $8.13$, and a CER \cite{cer} of $2.18$ on the Common Voice 8 \cite{commonvoice:2020} test set when combined with its language model decoder.
\subsubsection{OpenAI's Whisper}

OpenAI's Whisper \cite{whipser} architecture is a collection of multiple sizes of a single model designed to bridge the gap between small supervised models trained on limited data and large unsupervised models that require precise fine-tuning to perform specific tasks. For example, Whisper models were trained on speech recognition tasks, such as speech-to-text and language identification. Using a large amount of weakly-labelled data, the models can learn from various speech patterns and variations, resulting in improved robustness and generalization.

The training process for Whisper is scaled to $680$k hours of multilingual and multitask supervision, resulting in models that can generalize well to standard benchmarks and are often competitive with fully supervised models without the need for fine-tuning. For example, on the Common Voice \cite{commonvoice:2020} benchmark, Whisper models achieved a word error rate of $4.5\%$, which significantly improved over the previous state-of-the-art.

Of the $680$k hours of audio used for training, $117$k hours cover $96$ different languages, and the dataset also includes $125$k hours of language-to-English translation data. The model's capabilities and results on multiple datasets suggest that Whisper could be the next state-of-the-art in Swiss German ASR tasks. With access to a growing number of high-quality Swiss-German datasets, there is potential to create a new baseline for future Swiss German ASR tasks.

Additionally, as OpenAI's Whisper architecture includes model sizes ranging from tiny ($39$M parameters) to large ($1.5$B parameters), exploring how the different model sizes perform on the same tasks is an exciting way to explore.

\begin{figure*}[ht]
    \includegraphics[width=0.8\textwidth]{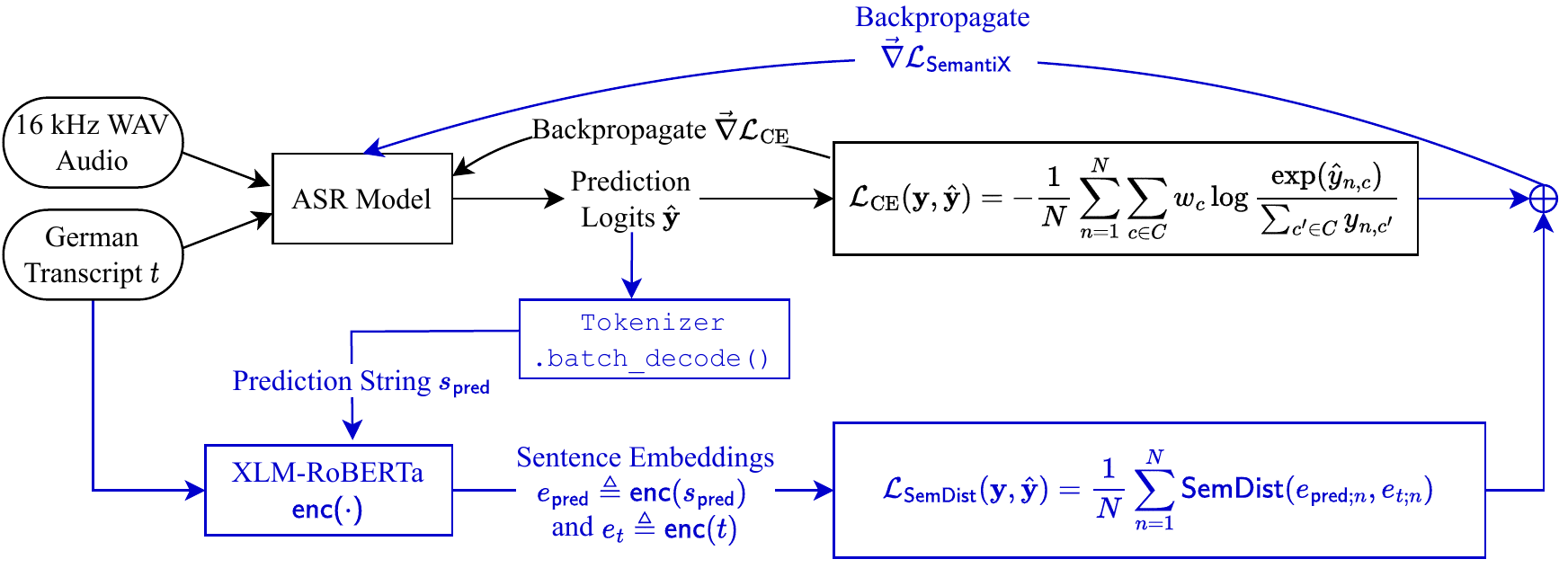}
    \centering
    \caption{Flowchart of training pipeline – the blue elements of the flowchart are needed for \lsem. The black elements are needed for training with the initial cross-entropy loss.}
\end{figure*}

\label{metrics}
\subsection{Inviting Semantics to Traditional ASR Metrics and Losses}

To evaluate our results, we use Word Error Rate (WER) \cite{wer} and Character Error Rate (CER) \cite{cer} metrics. Similarly to recent Swiss German ASR literature, we also compute the BLEU \cite{papineni-etal-2002-bleu} metric, which is standard in the automatic evaluation of machine translation. Indeed, our task is, to some extent, a translation task due to the particularities of Swiss German with respect to standard German, such as differences in verb conjugation, vocabulary or morphology. Examples of these differences can be found in the result section.


\subsubsection{Motivation to Involve Semantics}

Error-rate metrics are predominant metrics in the world of ASR. However, as described by Kim et al.\cite{sem-dist}, only using WER can be misleading, as it only takes into account literal correctness instead of semantic correctness. As shown in the paper, their proposed metric $\mathsf{SemDist}$ takes into account the semantics of the predicted sentence by leveraging the power of LLMs. It also inspired the creation of a new custom loss, \lsem, which we will describe in \ref{semantix-part}. Let us first introduce semantic distance as a metric and then explain how we used it to derive a new loss.

\subsubsection{Semantic Distance as a Metric}
Semantic distance uses the fact that LLMs can create word/sentence embeddings, which can be seen as point encoding in a latent semantic space. One can then compute the "distance" between those points using a score like cosine similarity. Facebook AI's paper by Kim et al. \cite{sem-dist} does precisely that by introducing $\mathsf{SemDist}$. It is defined as follows, with $\vec{x}$ and $\vec{y}$ two sentence embeddings encoded by a sentence encoder \cite{sent-encoder}.

\label{semdist}
\begin{equation}
    \mathsf{SemDist}(\vec{x}, \vec{y})  = 1 - \frac{ \vec{x}^\top \vec{y}}{||\vec{x}|| \cdot ||\vec{y}||}
\end{equation}
A result closer to zero indicates a higher semantic similarity and vice versa. We compute $\mathsf{SemDist}(\cdot, \cdot)$ during evaluation to add a level of comprehension to our results.

From this sample metric, we also derive a loss $\mathcal{L}_\mathsf{SD}$ on a batch of size $N$, with $\vec{y}=\{y_n\}_{n=1}^N$ and $\vec{\hat{y}}=\{\hat{y}_n\}_{n=1}^N$ respectively batches of labels and predictions as the mean of the respective distances:
\begin{equation}
    \mathcal{L}_\mathsf{SD}(\vec{y}, \hat{\vec{y}}) 
    \triangleq \frac{1}{N}\sum_{n=1}^{N} \mathsf{SemDist}\left\{\mathsf{enc}(y_n), \mathsf{enc}(\hat{y}_n)\right\}
\end{equation}

In our setup, $\mathsf{enc}(\cdot)$ corresponds to the sentence encoded by XLM-RoBERTa \cite{roberta} \cite{sent-encoder}.

\subsubsection{SemantiX Loss}
\label{semantix-part}

By default, Whisper uses a cross-entropy loss between the undecoded outputs of the model and the tokenized ground truth labels to quantify error. It is defined as follows, with $C$ the set of all possible tokens that the model can output:
\begin{equation}
    \mathcal{L}_\text{CE}(\mathbf{y}, \hat{\mathbf{y}}) = -\frac{1}{N} \sum_{n=1}^{N}\sum_{c\in C}{w_c} \log{\frac{\exp({\hat{y}_{n,c}})}{\sum_{c' \in C}{y_{n,c'}}}}
\end{equation}

We introduce a new loss \lsem, which uses both the well-known cross-entropy loss for literal correctness but also $\mathcal{L}_\mathsf{SD}$ for semantic correctness. The loss takes two hyperparameters $\alpha,\beta\in\mathbb{R}_+$. Combined, it gives the following function: 
\begin{equation}
    \label{semantix-sum}
    \mathcal{L}_\mathsf{SemantiX}(\vec{y}, \hat{\vec{y}}) \triangleq \alpha * \mathcal{L}_\mathsf{SD}(\vec{y}, \hat{\vec{y}}) + \beta * \mathcal{L}_{\mathsf{CE}}(\vec{y}, \hat{\vec{y}})
\end{equation}
It enables us to weigh the importance of the meaning of the prediction in addition to outputting the right tokens.
In our experiments, we also used a variant of it with a product instead of a sum:
\begin{equation}
    \label{semantix-prod}
    \mathcal{L}_\mathsf{SemantiX}'(\vec{y}, \hat{\vec{y}}) \triangleq \left(\gamma + \mathcal{L}_\mathsf{SD}(\vec{y}, \hat{\vec{y}}) \right) * \mathcal{L}_{\mathsf{CE}}(\vec{y}, \hat{\vec{y}})
\end{equation}
Both implementations use \textbf{\texttt{CosineEmbeddingLoss}} from PyTorch \cite{pytorch-paper}.

\subsection{Training Setup}
\subsubsection{XLS-R}
A Zero-Shot evaluation was first performed with the German pretrained XLS-R 1B model on the following datasets: SwissDial, SDS-200, SPC and Fleurs.
In a second time, we fine-tuned the model on the SwissDial and SDS-200 datasets, following the procedure suggested by the authors\footnote{\url{https://huggingface.co/blog/wav2vec2-with-ngram}}. We used the described hyper-parameters with a learning rate of $1\mathrm{e}-4$ and trained for $5$ epochs.

\subsubsection{Whisper}

Similarly to XLS-R 1B, we evaluated Whisper using a Zero-Shot approach on the following datasets: SwissDial, SDS-200, SPC and Fleurs. The first three datasets were used to baseline Whisper's initial performance on Swiss German. The last dataset was used as a control dataset to see how well our Whisper implementation's results aligned with Whisper's paper's results. Moreover, we evaluated all the different sizes of Whisper to compare them against each other.

Furthermore, we mostly fine-tuned the medium size of Whisper as it offered the best performance/training-time ratio. In addition, we fine-tuned Whisper on multiple combinations of three Swiss German datasets (see \ref{datasets}). As Whisper training time is consequent and we only fine-tuned it, we kept the number of training epochs close to one for each dataset. The different outcomes and conclusions are described in \ref{baselines}. Furthermore, we also fine-tuned Whisper using our innovative loss, described in \ref{metrics} (see Fig. \ref{post-training}). Our fine-tuning method was heavily inspired by the procedure described by Sanchit Ghandi\footnote{\url{https://huggingface.co/blog/fine-tune-whisper}}.

\label{results}

\section{Results}

\begin{table*}[ht]
\begin{center}
    \small
    \begin{tabular}{p{0.3\linewidth}|p{0.3\linewidth}|l|l|l|l}
    \toprule
    \textbf{Ground Truth} & \textbf{Prediction} &
    \textbf{WER} & \textbf{CER} & \textbf{BLEU} & \textbf{SemDist} \\
    \midrule
    Andererseits seien strategische Entscheide für den Rückgang verantwortlich.
    &
    Andererseits sei ein strategischer Entscheid für den Rückgang verantwortlich.
    & $50.0$ & $5.3$
    & $41.1$ & $1.0$ \\
    \midrule
    Boeing lehnte eine Stellungnahme ab.
    &
    Boeing hat den Stellungnahme abgelehnt.
    & $60.0$ & $38.9$
    & $0.0$ & $1.1$ \\
    \midrule
    Wegen des Brandes war die Dorf\textbf{strasse} für mehrere Stunden gesperrt.
    &
    Wegen des Brandes war die Dorf\textbf{strafe} mehrere Stunden gesperrt.
    & $20.0$ & $10.0$
    & $59.5$ & $23.3$ \\
    \midrule
    Aus Syrien stammten im Mai 52 Asyl\textbf{bewerber}.
    &
    Aus Syrien stammten im Mai 52 Asyl\textbf{werber}.
    & $14.3$ & $4.7$
    & $70.7$ & $0.2$ \\
    \midrule
    Inzwischen ist es kurz vor 22 Uhr.
    &
    Mittlerweile ist es kurz vor 10 Uhr.	
    & $28.6$ & $38.2$
    & $41.1$ & $23.4$ \\
    \bottomrule
    \end{tabular}
    \caption{Selected examples of predictions. It highlights the quality of $\mathsf{SemDist}$ as a metric \label{example_table}}
\end{center}
\end{table*}

\subsection{Baselines}
\label{baselines}

\begin{figure}[ht]
    \centering
    \includegraphics[width=0.5\textwidth]{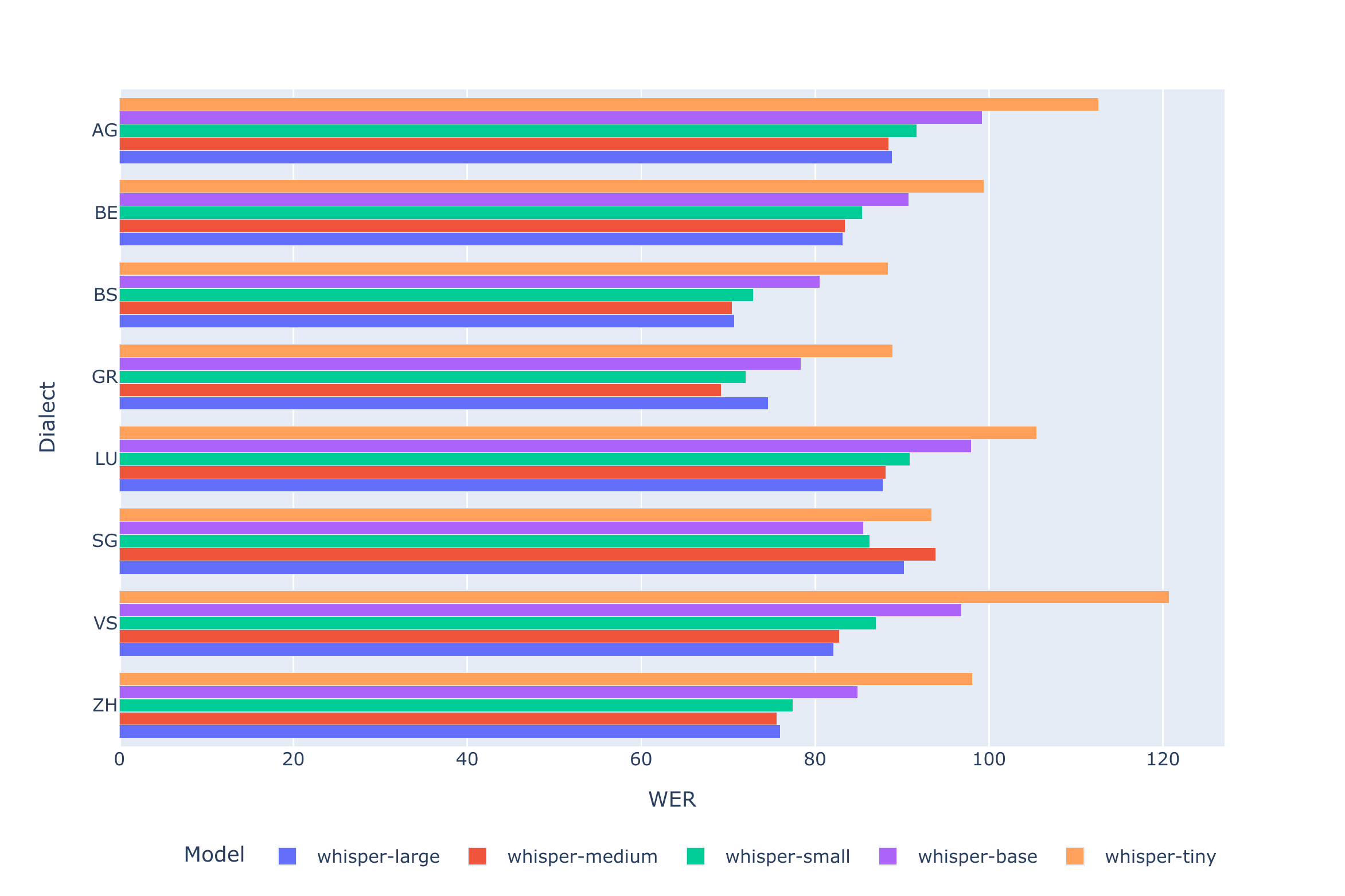}
    \caption{Zero-Shot evaluation results of different Whisper sizes on the SwissDial dataset, with dialectal transcripts, \label{plots}}
\end{figure}

Our first experiment was a Zero-shot evaluation on the SwissDial test set using the dialectal transcripts. As shown in Figure \ref{plots}, the results are quite bad. This can be explained by the lack of a standardized written form of the Swiss German language and the fact that Whisper has not been introduced to any sort of Swiss German transcript, which renders it close to impossible to produce a correct output. These results and a secondary evaluation with standard German transcripts reported in Table \ref{zeroshot} confirmed our decision to use those for the rest of the experiments.

In a second time, we evaluated XLS-R and every size of Whisper, in a similar Zero-Shot \cite{Zero-Shot} fashion, on each of our datasets with standard German transcripts. We include the Fleurs dataset to assess how well each of the model performs on standard German. All these results are reported in Table \ref{zeroshot}. Even though Whisper Large performs slightly better on SDS-200 and Fleurs, we decided to focus on Whisper Medium model for further fine-tuning and for computational reasons. A huge improvement is still noticeable between different sizes of Whisper, and XLS-R seems to perform worse on Zero-Shot evaluation. The well known ability of Whisper to generalize to unknown data and regional accents might explain these results.

\label{our-models}
\subsection{Our Models}
Based on the baseline results, we decided to fine-tune Whisper Medium on our datasets, leveraging the semantic losses (Table \ref{post-training}). \wsem has been trained first on a shuffled concatenation of SwissDial and $20\%$ of SPC (to have roughly as many samples from both) using cross-entropy loss for $1$ epoch over the whole train dataset ($\approx60$k samples), then trained for another round on SDS-200 for $2$ epochs on the full train dataset ($\approx37$k samples), this time using \lsem. \wce has been trained in the same setting but sticking to $\mathcal{L}_\text{CE}$ the whole training. Both models were trained with a batch size of $10$ and $2$ gradient accumulation steps – depending on the runs, on NVIDIA A$100$ and RTX Quadro $6000$ GPUs. In this setup, we obtain better results than state-of-the-art on SDS-200 with \wce for CER and with \wsem for BLEU and WER.

We observe that after training Whisper on any of our datasets, it tends to suffer from catastrophic forgetting, as we see, for instance, in the performance decrease for the Fleurs dataset, between fine-tuned models results and the Zero-Shot ones. To avoid this, the models should probably have been trained on all the datasets altogether for $2$ to $3$ epochs in a similar fashion as in the Whisper paper \cite{whipser}.

\begin{table*}[ht]
    \small

\begin{center}
    \begin{tabular}{l|c|c|c}
    \toprule
    & \textbf{Whisper Medium} & \textbf{\wce} & \textbf{\wsem} \\
    \midrule
    \textbf{SwissDial} & $9.2\%$ & $5.0\%$ & $4.4\%$ \\
    \textbf{SDS-200}   & $11.0\%$ & $4.9\%$ & $4.2\%$ \\
    \textbf{SPC}       & $8.8\%$ & $8.1\%$ & $6.7\%$ \\
    \textbf{Fleurs}    & $2.5\%$ & $5.7\%$ & $5.3\%$ \\
    \bottomrule
    \end{tabular}
    \caption{Average semantic distance $\mathcal{L}_\mathsf{SemDist}$, as a metric, during evaluation on our fine-tuned models \label{sd-table-transposed}}
\end{center}
\end{table*}

Note that Whisper – as opposed to XLS-R 1B – works with unnormalized text – namely, it preserves punctuation, casing, digits, onomatopoeia, etc – for both input and output. To our understanding, evaluation metrics in Swiss German ASR literature, namely WER and BLEU, were computed in a normalized setting. The transcription contained the characters \texttt{a-z}, \texttt{\"a}, \texttt{\"o}, \texttt{\"u}, and spaces, with no punctuation, casing, and numbers spelt out. We computed our metrics on normalized predictions and ground truth but observed only minor differences, suggesting that Whisper interpolates punctuation and casing very well. In some examples presented in Table \ref{example_table}, we observe that Whisper encounters difficulties in transposing verb tenses from the Swiss German conjugation to Standard German. Probably due to such differences – WER results might be too pessimistic. The 2nd sentence is a good example, where we observe a high WER but a low $\mathsf{SemDist}$, because the meaning of the predicted sentence remains very close. Therefore, we find that even though WER might seem discouraging, the overall $\mathcal{L}_\mathsf{SemDist}$ is very low on our models for all of our datasets (Fig. \ref{sd-table-transposed}) and suggests good transcription capacities by our model. We performed a $p$-value test between the standard metrics and $\mathsf{SemDist}$ on our model predictions and observed no significant correlation. Hence, $\mathsf{SemDist}$ seems to be an interesting new metric that could bring a new level of understanding in ASR transcription tasks. \\
We believe that semantic distance could significantly improve ASR task evaluation when combined with error-rate measures. Additional examples can be found in Table \ref{example_table}.

\section{Discussion}


Using the large version of Whisper for Zero-Shot evaluation on our datasets also showed promising results. Therefore, one should try our experiments with the large version of Whisper if one has the computing capacities. 

Nevertheless, an observed downside of the Whisper fine-tuning is the catastrophic forgetting of our model. Indeed, after training, one can see that the models perform worse even in German. We believe however that, with enough computing capacities, this catastrophic forgetting could be attenuated while conserving the encouraging training results presented in this paper.

Another important finding of this study is that traditional metrics might not be sufficiently extensive to capture all the intrinsics of the "Swiss German audio to High German text" task. The shortcomings of traditional metrics and losses opened the possibility to have predicted sentences to be further processed, using for instance a LLM to match the conjugation rules.

\section*{Summary}

This study shows that the newly released Whisper model is able to impressively generalize its knowledge to unseen languages such as Swiss German dialects, despite important disparities between speech and transcription. Moreover, fine-tuning Whisper on a diverse set of Swiss German datasets can significantly improve its overall performance. As presented before, we were able to fine-tune Whisper and outperform state-of-the-art using SwissDial, SDS-200, and to a limited extent SPC. These results were obtained with minimal hyperparameters tuning and a small number of epochs. We used $\mathsf{SemDist}$ to offer additional insight into our results and to define a custom loss function \lsem for training. However, we did not get consistently better results in traditional metrics such as WER, BLEU and CER. Nevertheless, our custom loss function helped reduce the semantic distance between our predictions and transcripts on Swiss German datasets.

\newpage

\bibliography{bibi}
\bibliographystyle{acl_natbib}

\end{document}